\documentclass[twocolumn]{ceurart}

\usepackage{listings}
\usepackage[a4paper]{geometry} 
\usepackage{graphicx}
\usepackage{xurl} 
\usepackage[italian,english]{babel}
\usepackage{latexsym} 
\usepackage{xcolor}
\usepackage{verbatim}
\usepackage{amsmath}

\usepackage{tabularx}
\usepackage{diagbox}
\usepackage{appendix}
\usepackage{enumitem}

\newcommand{\takeout}[1]{}

\newcommand{\sg}[1]{\textcolor{blue}{#1-sg}}
\newcommand{\pl}[1]{\textcolor{red}{#1-pl}}
\newcommand{\ul}{\underline}
\begin{document}

\copyrightyear{2024}
\copyrightclause{Copyright for this paper by its authors.
  Use permitted under Creative Commons License Attribution 4.0
  International (CC BY 4.0).}

\conference{CLiC-it 2024: Tenth Italian Conference on Computational Linguistics, Dec 04 — 06, 2024, Pisa, Italy}

\title{Exploring syntactic information in sentence embeddings through multilingual subject-verb agreement 
}

\author[1]{Vivi Nastase}[email=vivi.a.nastase@gmail.com]
\cormark[1]
\author[1,2]{Chunyang Jiang}[email=chunyang.jiang42@gmail.com]
\author[1]{Giuseppe Samo}[email=giuseppe.samo@idiap.ch]
\author[1,2]{Paola Merlo}[email=Paola.Merlo@unige.ch]
\address[1]{Idiap Research Institute, Martigny, Switzerland}
\address[2]{University of Geneva, Geneva, Switzerland}

\cortext[1]{Corresponding author.}


\begin{abstract}
In this paper, our goal is to investigate to what degree multilingual pretrained language models capture cross-linguistically valid abstract linguistic representations. We take the approach of developing curated synthetic data on a large scale, with specific properties, and using them to study sentence representations built using pretrained language models. We use a new multiple-choice task and datasets, Blackbird Language Matrices (BLMs), to focus on a specific grammatical structural phenomenon -- subject-verb agreement across a variety of sentence structures -- in several languages. Finding a solution to this task requires a system detecting complex linguistic patterns and paradigms in text representations. Using a two-level architecture that solves the problem in two steps -- detect syntactic objects and their properties in individual sentences, and find patterns across an input sequence of sentences -- we show that despite having been trained on multilingual texts in a consistent manner, multilingual pretrained language models have language-specific differences, and syntactic structure is not shared, even across closely related languages. \\ 

\noindent Questo lavoro chiede se i modelli linguistici multilingue preaddestrati catturino rappresentazioni linguistiche astratte valide attraverso svariate lingue. Il nostro approccio sviluppa dati sintetici curati su larga scala, con proprietà specifiche, e li utilizza per studiare le rappresentazioni di frasi costruite con modelli linguistici preaddestrati. Utilizziamo un nuovo \textit{task} a scelta multipla e i dati afferenti, le \textit{Blackbird Language Matrices} (BLM), per concentrarci su uno specifico fenomeno strutturale grammaticale - l'accordo tra il soggetto e il verbo  - in diverse lingue. Per trovare la soluzione corretta a questo \textit{task} è necessario un sistema che rilevi modelli e paradigmi linguistici complessi nelle rappresentazioni testuali. Utilizzando un'architettura a due livelli che risolve il problema in due fasi - prima impara gli oggetti sintattici e le loro proprietà nelle singole frasi e poi ne ricava gli elementi comuni - dimostriamo che, nonostante siano stati addestrati su testi multilingue in modo coerente, i modelli linguistici multilingue preaddestrati presentano differenze specifiche per ogni lingua e inoltre la struttura sintattica non è condivisa, nemmeno tra lingue tipologicamente molto vicine. \\

\end{abstract}


\maketitle


\section{Introduction}

Large language models, trained on huge amount of texts, have reached a level of performance that rivals human capabilities on a range of established benchmarks \cite{wang_superglue}. Despite high performance on high-level language processing tasks, it is not yet clear what kind of information these language models encode, and how. For example, transformer-based pretrained models have shown excellent performance in tasks that seem to require that the model encodes syntactic information \cite{Manning2020EmergentLS}. 

All the knowledge that the LLMs encode comes from unstructured texts and the shallow regularities they are very good at detecting, and which they are able to leverage into information that correlates to higher structures in language. Most notably, \cite{rogers-etal-2020-primer} have shown that from the unstructured textual input, BERT \citep{devlin-etal-2019-bert} is able to infer POS, structural, entity-related, syntactic and semantic information at successively higher layers of the architecture, mirroring the classical NLP pipeline \citep{tenney-etal-2019-bert}. We ask: How is this information encoded in the output layer of the model, i.e. the embeddings? Does it rely on surface information -- such as inflections, function words -- and is assembled on the demands of the task/probes \citep{hewitt-liang-2019-designing}, or does it indeed reflect something deeper that the language model has assembled through the progressive transformation of the input through its many layers? 

To investigate this question, we use a seemingly simple task -- subject-verb agreement. Subject-verb agreement is often used to test the syntactic abilities of deep neural networks \cite{linzen2016,gulordava2018,goldberg2019,linzen2021}, because, while apparently simple and linear, it is in fact structurally, and theoretically, complex, and requires connecting the subject and the verb across arbitrarily long or complex structural distance. It has an added useful dimension -- it relies on syntactic structure and grammatical number information that many languages share.

In previous work we have shown that simple structural information -- the chunk structure of a sentence -- which can be leveraged to determine subject-verb agreement, or to contribute towards more semantic tasks, can be detected in the sentence embeddings obtained from a pre-trained model \cite{nastase2024identifiable}. This result, though, does not cast light on whether the discovered structure is deeper and more abstract, or it is rather just a reflection of surface indicators, such as function words or morphological markers. 

To tease apart these two options, we set up an experiment covering four languages: English, French, Italian and Romanian. These languages, while different, have shared properties that make sharing of syntactic structure a reasonable expectation, if the pretrained multilingual model does indeed discover and encode syntactic structure. We use parallel datasets in the four languages, built by (approximately) translating the BLM-AgrF dataset \cite{an-etal-2023-blm}, a multiple-choice linguistic test inspired from the Raven Progressive Matrices visual intelligence test, previously used to explore subject-verb agreement in French. 

Our work offers two contributions: (i) four parallel datasets -- on English, French, Italian and Romanian, focused on subject-verb agreement; (ii) cross-lingual and multilingual testing of a multilingual pretrained model, to explore the degree to which syntactic structure information is shared across different languages. Our cross-lingual and multilingual experiments show poor transfer across languages, even those most related, like Italian and French. This result indicates that pretrained models encode syntactic information based on shallow and language-specific clues, from which they are not yet able to take the step towards abstracting grammatical structure.
The datasets are available at \url{https://www.idiap.ch/dataset/(blm-agre|blm-agrf|blm-agri|blm_agrr)} and the code at \url{https://github.com/CLCL-Geneva/BLM-SNFDisentangling}.

\section{BLM task and BLM-Agr datasets}
\label{sec:data}


Inspired by existing IQ tests ---Raven's progressive matrices (RPMs)--- we have developed a framework, called Blackbird Language Matrices (BLMs) \cite{merlo2023} and several datasets \cite{an-etal-2023-blm,samo-etal-2023}.
RPMs consist of a sequence of images, called the \textit{context}, connected in a logical sequence by underlying generative rules \cite{raven1938}. The task is to determine the  missing element in this visual sequence, the \textit{answer}. The candidate answers are constructed to be similar enough that the solution can be found only if the rules are identified correctly. 

\begin{figure}
\begin{flushleft}
    \small
    \begin{tabular}{lllll} 
    \hline
    \multicolumn{5}{c}{\sc Context}\\
    \hline
    1 & \sg{NP}& \sg{PP1}& & \sg{VP}  \\
    2 & \pl{NP} & \sg{PP1}& & \pl{VP}  \\
    3 & \sg{NP}& \pl{PP1} & & \sg{VP}  \\
    4 & \pl{NP} & \pl{PP1} & & \pl{VP}  \\
    5 & \sg{NP}& \sg{PP1}& \sg{PP2} &\sg{VP}  \\
    6 & \pl{NP} & \sg{PP1}  & \sg{PP2} & \pl{VP}  \\
    7 & \sg{NP}   & \pl{PP1} & \sg{PP2} &  \sg{VP}  \\
    8 & ??? & & & \\ \hline
    \end{tabular}
    
    \begin{tabular}{lllllr} \hline
    \multicolumn{6}{c}{{\sc Answers} } \\ \hline
    \rowcolor{lightgray!40} 
    1 & \pl{NP} & \pl{PP1} & \sg{PP2} & \pl{VP} & \textsc{Correct} \\
    2 & \pl{NP} & \pl{PP1} & et \sg{PP2} & \pl{VP} & Coord  \\ 
    3 & \pl{NP} & \pl{PP1} &          & \pl{VP} & WNA\\
    4 & \pl{NP} & \sg{PP1} & \sg{PP1} & \pl{VP} & WN1 \\
    5 & \pl{NP} & \pl{PP1} & \pl{PP2} & \pl{VP} & WN2 \\
    6 & \pl{NP} & \pl{PP1} & \pl{PP2} & \sg{VP} & AEV \\
    7 & \pl{NP} & \sg{PP1} & \pl{PP2} & \sg{VP} & AEN1 \\
    8 & \pl{NP} & \pl{PP1} & \sg{PP2} & \sg{VP} & AEN2 \\ \hline
     \end{tabular}
\end{flushleft}

\caption{BLM instances for verb-subject agreement, with two attractors. The errors can be grouped in two types: (i) \ul{sequence errors}: WNA= wrong nr. of attractors; WN1= wrong gram. nr. for 1$^{st}$ attractor noun (N1); WN2= wrong gram. nr. for 2$^{nd}$ attractor noun (N2); (ii) \ul{grammatical errors}: AEV=agreement error on the verb; AEN1=agreement error on N1; AEN2=agreement error on N2.}
\label{fig:matrices}
\vspace{-5mm}
\end{figure}

Solving an RPM problem is usually done in two steps: (i) identify the relevant objects and their attributes; (ii) decompose the main problem into subproblems, based on object and attribute identification, in a way that allows detecting the global pattern or underlying rules \cite{carpenter1990one}. 

Such an approach can be very useful for probing language models, as it allows to test whether they indeed detect the relevant linguistic objects and their properties, and whether (or to what degree) they use this information to find larger patterns. We have developed BLMs as a linguistic test.  Figure \ref{fig:matrices} illustrates the template of a BLM subject-verb agreement matrix, with the different linguistic objects -- chunks/phrases -- and their relevant properties, in this case grammatical number. Examples in all languages under investigation are provided in Appendix~\ref{app:agr}. 

\paragraph{BLM-Agr datasets}
\label{sec:blmdata}

A BLM problem for subject-verb agreement consists of a context set of seven sentences that share the subject-verb agreement phenomenon, but differ in other aspects -- e.g. number of linearly intervening noun phrases between the subject and the verb (called attractors because they can interfere with the agreement), different grammatical numbers for these attractors, and different clause structures. The sequence is generated by a rule of progression of number of attractors, and alternation in the grammatical number of the different phrases. Each context is paired with a set of candidate answers generated from the correct answer by altering it to produce minimally contrastive error types. We have two types of errors (see Figure \ref{fig:matrices}: (i) sequence errors -- these candidate answers are grammatically correct, but they are not the correct continuation of the sequence; (ii) agreement errors -- these candidate answers are grammatically erroneous, because the verb is in agreement with one of the intervening attractors. By constructing candidate answers with such specific error types, we can investigate the kind of information and structure learned.

The seed data for French was created by manually completing data previously published data \cite{franck2002subject}. From this initial data, we generated a dataset that comprises three subsets of increasing lexical complexity (details in \cite{an-etal-2023-blm}): Types I, II, III,  corresponding to different amounts of lexical variation within a problem instance. Each subset contains three clause structures uniformly distributed within the data. The dataset used here is a variation of the BLM-AgrF \citep{an-etal-2023-blm} that separates sequence-based from other types of errors, to be able to perform deeper analyses into the behaviour of pretrained language models. 

The datasets in English, Italian and Romanian were created by manually translating the seed French sentences into the other languages by native (Italian and Romanian) and near-native (English) speakers. The internal structure in these languages is very similar, so translations are approximately parallel. The differences lie in the treatment of preposition and determiner sequences that must be conflated into one word in some cases in Italian and French, but not in English. 
French and Italian use number-specific determiners and inflections, while Romanian and English encode grammatical number exclusively through inflections. In English most plural forms are marked by a suffix. Romanian has more variation, and noun inflections also encode case. Determiners are separate tokens, which are overt indicators of grammatical number and of phrase boundaries, whereas inflections may or may not be tokenized separately. 

Table \ref{tab:multilingual-stats} shows the datasets statistics for the four BLM problems. After splitting each subset 90:10 into train:test subsets, we randomly sample 2000 instances as train data. 20\% of the train data is used for development. 

\begin{table}[h]
\setlength{\tabcolsep}{2mm}
\small
\begin{tabular}{l|c|c|c|c}
& \textbf{English} & \textbf{French} & \textbf{Italian} &\textbf{Romanian} \\ \hline
Type I  & 230 &  252 & 230 & 230 \\
Type II & 4052 & 4927  & 4121 & 4571 \\
Type III & 4052 & 4810 & 4121 & 4571 \\ \hline
\end{tabular}
\caption{Test data statistics. The amount of training data is always 2000 instances.}
\label{tab:multilingual-stats}
\vspace{-5mm}
\end{table}

\paragraph{A sentence dataset}
\label{sec:sentencedata}

From the seed files for each language we build a dataset to study sentence structure independently of a task. The seed files contain noun, verb and prepositional phrases, with singular and plural variations. From these chunks, we build sentences with all (grammatically correct) combinations of \texttt{np [pp$_1$ [pp$_2$]] vp}\footnote{pp$_1$ and pp$_2$ may be included or not, pp$_2$ may be included only if pp1 is included}. For each chunk pattern $p$ of the 14 possibilities (e.g.,  $p$ = "np-s pp1-s vp-s"), all corresponding sentences are collected into a set $S_p$. 

The dataset consists of triples $(in, out^+, Out^-)$, where $in$ is an input sentence, $out^+$ is the correct output -- a sentence different from $in$ but with the same chunk pattern. $Out^-$ are $N_{negs}=7$ incorrect outputs, randomly chosen from the sentences that have a chunk pattern different from $in$. For each language, we sample uniformly approx. 4000 instances from the generated data based on the pattern of the input sentence, randomly split 80:20 into train:test. The train part is split 80:20 into train:dev, resulting in a 2576:630:798 split for train:dev:test.

\section{Probing the encoding of syntax}

We aim to test whether the syntactic information detected in multilingual pretrained sentence embeddings is based on shallow, language-specific clues, or whether it is more abstract structural  information. Using the subject-verb agreement task and the parallel datasets in four languages provides clues to the answer. 

The datasets all share sentences with the same syntactic structures, as illustrated in Figure \ref{fig:matrices}. However, there are language specific differences, as in the structure of the chunks (noun or verb or prepositional phrases) and each language has different ways to encode grammatical number (see section \ref{sec:blmdata}).

If the grammatical information in the sentences in our dataset -- i.e. the sequences of chunks with specific properties relevant to the subject-verb agreement task (Figure \ref{fig:matrices}) -- is an abstract form of knowledge within the pretrained model, it will be shared across languages. We would then see a high level of performance for a model trained on one of these languages, and tested on any of the other. Additionally, when training on a dataset consisting of data in the four languages, the model should detect a shared parameter space that would lead to high results when testing on data for each language.

If however the grammatical information is a reflection of shallow language indicators, we expect to see higher performance on languages that have overt grammatical number and chunk indicators, such as French and Italian, and a low rate of cross-language transfer.

\subsection{System architectures}
\label{sec:parts-for-BLMs}

\paragraph{A sentence-level VAE}

To test whether chunk structure can be detected in sentence embeddings we use a VAE-like system, which encodes a sentence, and decodes a different sentence with the same chunk structure, using a set of contrastive negative examples -- sentences that have different chunk structures from the input -- to encourage the latent to encode the chunk structure.

The architecture of the sentence-level VAE is similar to a previously proposed system \cite{nastase-merlo-2023-grammatical}: the encoder consists of a CNN layer with a 15x15 kernel, which is applied to a 32x24-shaped sentence embedding, followed by a linear layer that compresses the output of the CNN into a latent layer of size 5. The decoder mirrors the encoder.

An instance consists of a triple $(in, out^+, Out^-)$, where $in$ is an input sentence with embedding $e_{in}$ and chunk structure $p$, $out^+$ is a sentence with embedding $e_{out^+}$ with  same chunk structure $p$, and $Out^- = \{s_k| k=1,N_{negs}\}$ is a set of $N_{negs}=7$ sentences with embeddings $e_{s_k}$, each with chunk pattern different from $p$ (and different from each other). The input $e_{in}$ is encoded into latent representation $z_i$, from which we sample a vector $\tilde{z}_i$, which is decoded into the output $\hat{e}_{in}$. To encourage the latent to encode the structure of the input sentence we use a max-margin loss function, to push for a higher similarity score for $\hat{e}_{in}$ with the sentence that has the same chunk pattern as the input ($e_{out^+}$) than the ones that do not. At prediction time, the sentence from the $\{out^+\} \cup Out^-$ options that has the highest score relative to the decoded answer is taken as correct.

\paragraph{Two-level VAE for BLMs}

\begin{figure}[h]
    \centering
    \includegraphics[width=0.49\textwidth]{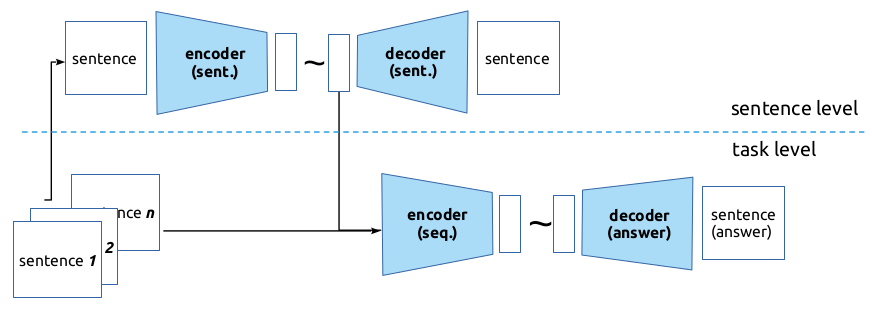}
    \caption{A two-level VAE: the sentence level learns to compress a sentence into a representation useful to solve the BLM problem on the task level.}
    \label{fig:2levelVAE}
    \vspace{-3mm}
\end{figure}

We use a two-level system illustrated in Figure \ref{fig:2levelVAE}, which separates the solving of the BLM task on subject-verb agreement into two steps: (i) compress sentence embeddings into a representation that captures the sentence chunk structure and the relevant chunk properties (on the sentence level) (ii) use the compressed sentence representations to solve the BLM agreement problems, by detecting the pattern across the sequence of structures (on the task level). This architecture will allow us to test whether sentence structure -- in terms of chunks -- is shared across languages in a pretrained multilingual model. 

All reported experiments use Electra \citep{clark2020electra}\footnote{Electra pretrained model: google/electra-base-discriminator}, with the sentence representations the embedding of the [CLS] token (details in \cite{nastase2024identifiable}).

An instance for a BLM problem consists of an ordered context sequence $S$ of sentences, $S = \{s_i| i = 1,7\}$ as input, and an answer set $A$ with one correct answer $a_c$, and several incorrect answers $a_{err}$. Every sentence is embedded using the pretrained model. To simplify the discussion, in the sections that follows, when we say {\it sentence} we actually mean its embedding.

The two-level VAE system takes a BLM instance as input, decomposes its context sequence $S$ into sentences and passes them  individually as input to the sentence-level VAE. For each sentence $s_i \in S$, the system builds on-the-fly the candidate answers for the sentence level: the same sentence $s_i$ from input is used as the correct output, and a random selection of sentences from $S$ are the negative answers. After an instance is processed by the sentence level, for each sentence $s_i \in S$, we obtain its representation from the latent layer $l_{s_i}$, and reassemble the input sequence as $S_l = stack[l_{s_i}]$, and pass it as input to the task-level VAE. The loss function combines the losses on the two levels -- a max-margin loss on the sentence level that contrasts the sentence reconstructed on the sentence level with the correct answer and the erroneous ones, and a max-margin loss on the task level that contrasts the answer constructed by the decoder with the answer set of the BLM instance (details in \cite{nastase2024identifiable}).

\subsection{Experiments}

To explore how syntactic information -- in particular chunk structure -- is encoded, we perform cross-language and multi-language experiments, using first the sentences dataset, and then the BLM agreement task. We report F1 averages over three runs.

Cross-lingual experiments -- train on data from one language, test on all the others -- show whether patterns detected in sentence embeddings that encode chunk structure are transferable across languages. The results on testing on the same language as the training provide support for the experimental set-up -- the high results show that the pretrained language model used does encode the necessary information, and the system architecture is adequate to distill it.

The multilingual experiments, where we learn a model from data in all the languages, will provide additional clues -- if the performance on testing on individual languages is comparable to when training on each language alone, it means some information is shared across languages and can be beneficial. 


\subsubsection{Syntactic structure in sentences}
We use only the sentence level of the system illustrated in Figure \ref{fig:2levelVAE} to explore chunk structure in sentences, using the data described in Section \ref{sec:sentencedata}. For the cross-lingual experiments, the training dataset for each language is used to train a model that is then tested on each test set. For the multilingual setup, we assemble a common training data from the training data for all languages. 
\subsubsection{Solving the BLM agreement task}
We solve the BLM agreement task using the two-level system, where a compacted sentence representation learned on the sentence level should help detect patterns in the input sequence of a BLM instance. Because the datasets are parallel, with shared sentence and sequence patterns, we test whether the added learning signal from the task level can help push the system to learn to map an input sentence into a representation that captures structure shared across languages. We perform cross-lingual experiments, where a model is trained on data from one language, and tested on all the test sets, and a multilingual experiment, where for each type I/II/III data, we assemble a training dataset from the training sets of the same type from the other languages. The model is then tested on the separate test sets.

\subsection{Evaluation}

For each training set we build three models, and plot the average F1 score. The standard deviation is very small, so we do not include it in the plot, but it is reported in the results Tables in Appendix \ref{app:results}.

\section{Results}

\begin{figure}[h]
    \centering
    \includegraphics[width=0.5\textwidth]{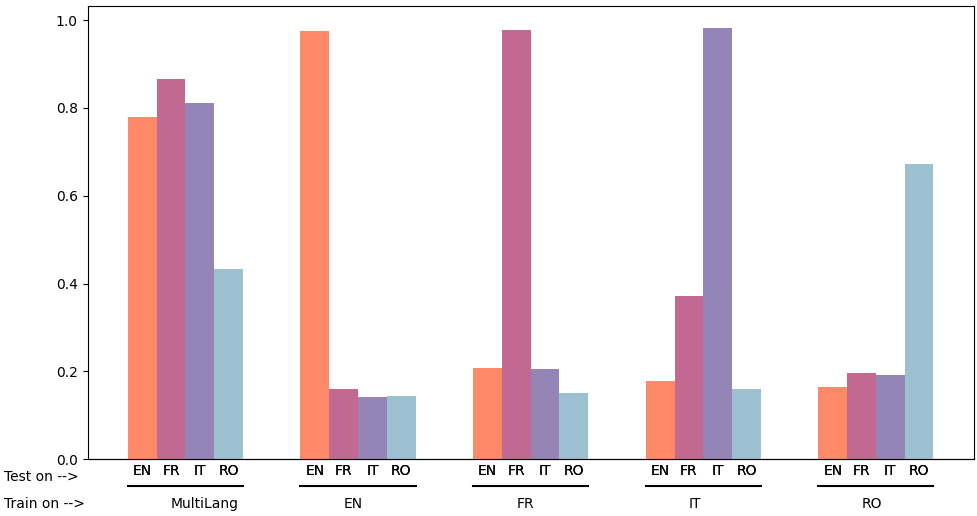}
    \caption{Cross-language testing for detecting chunk structure in sentence embeddings.}
    \label{fig:sentenceresults_Xlang}
    \vspace{-3mm}
\end{figure}

\paragraph{Structure in sentences}
Figure  \ref{fig:sentenceresults_Xlang} shows the results for the experiments on detecting chunk structure in sentence embeddings, in cross-lingual and multilingual training setups, for comparison (detailed results in Table \ref{tab:sentdet}). 

Two observations are relevant to our investigation: (i) while training and testing on the same language leads to good performance -- indicating that Electra sentence embeddings do contain relevant information about chunks, and that the system does detect the chunk pattern in these representations -- there is very little transfer effect. A slight effect is detected for the model learned on Italian and tested on French; (ii) learning using multilingual training data leads to a deterioration of the performance, compared to learning in a monolingual setting. This again indicates that the system could not detect a shared parameter space for the information that is being learned, the chunk structure, and thus this information is encoded differently in the languages under study.

\begin{figure}[h]
    \centering
    \includegraphics[width=0.45\textwidth,trim={2cm 1cm 2cm 2.5cm},clip]{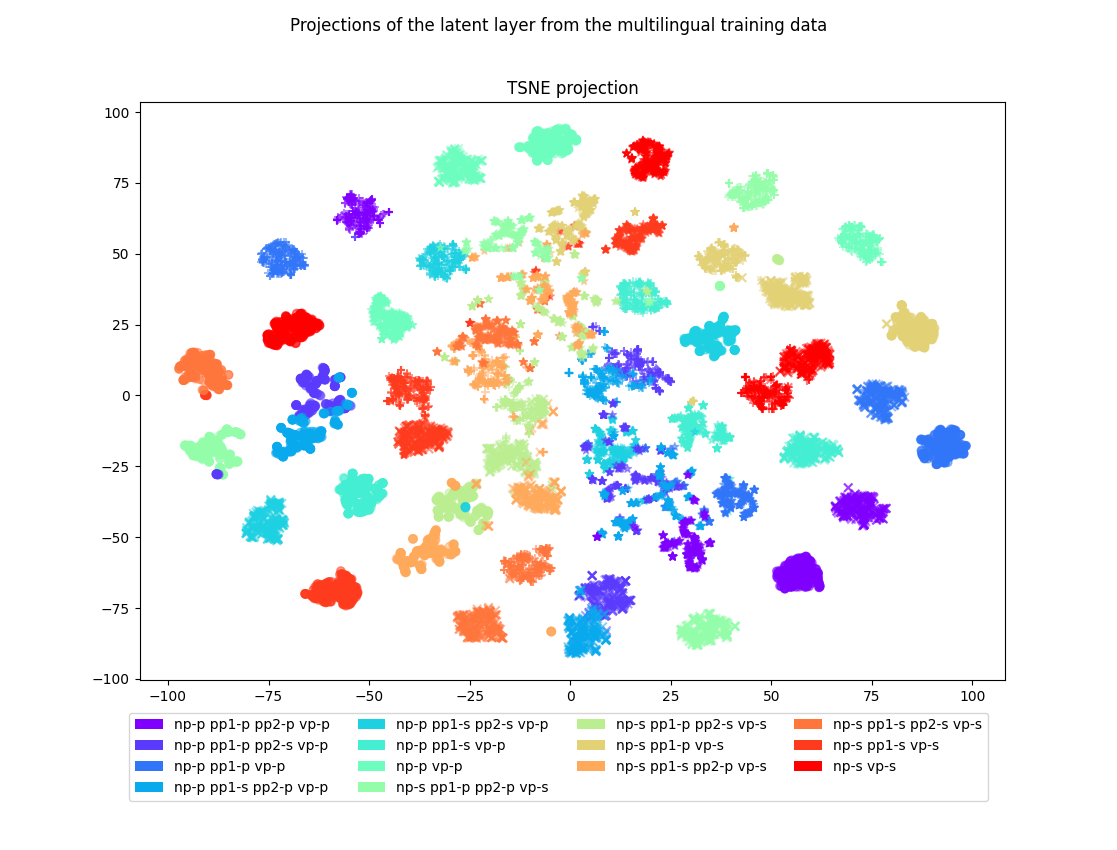}
    \caption{tSNE projection of the latent representation of sentences from the training data, coloured by their chunk pattern. Different markers indicate the languages: "o" for English, "x" for French, "+" for Italian, "*" for Romanian. We note that while representations cluster by the pattern, the clusters for different languages are disjoint.}
    \label{fig:sentplot}
\end{figure}

An additional interesting insight comes from the analysis of the latent layer representations. Figure \ref{fig:sentplot} shows the tSNE projection of the latent representations of the sentences in the training data after multilingual training. Different colours show different chunk patterns, and different markers show different languages. Had the information encoding syntactic structure been shared, the clusters for the same pattern  in the different languages would overlap. Instead, we note that each language seems to have its own quite separate pattern clusters.

\paragraph{Structure in sentences for the BLM agreement task}

\begin{figure}[h]
    \centering
    \includegraphics[width=0.5\textwidth,trim={0.5cm 0cm 0cm 0cm},clip]{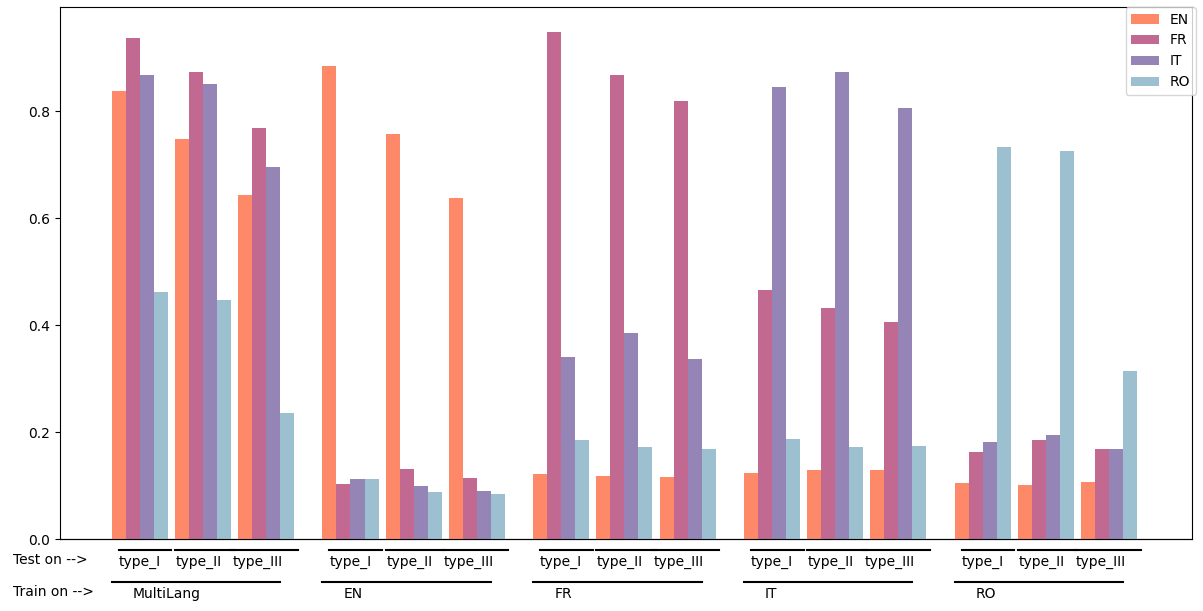}
    \caption{Average F1 performance on training on type I data over three runs -- cross-language and multi-language}
    \label{fig:BLMbars}
    \vspace{-7mm}
\end{figure}

When the sentence structure detection is embedded in the system for solving the BLM agreement task, where an additional supervision signals comes from the task, we note a similar result as when processing the sentences individually. Figure \ref{fig:BLMbars} shows the results for the multi-lingual and monolingual training setups for the type I data. Complete results are in Tables \ref{tab:multilingualBLM}-\ref{tab:monolingualBLM} in the appendix. 

\paragraph{Discussion and related work}

Pretrained language models are learned from shallow cooccurrences through a lexical prediction task. The input information is transformed through several transformer layers, various parts boosting each other through self-attention. Analysis of the architecture of transformer models, like BERT \cite{devlin-etal-2019-bert}, have localised and followed the flow of specific types of linguistic information through the system \citep{tenney-etal-2019-learn-from-context,rogers-etal-2020-primer}, to the degree that the classical NLP pipeline seems to be reflected in the succession of the model's layers. Analysis of contextualized token embeddings shows that they can encode specific linguistic information, such as sentence structure \citep{hewitt-manning-2019-structural} (including in a multilingual set-up \cite{chi-etal-2020-finding}), predicate argument structure \cite{conia-etal-2022-semantic}, subjecthood and objecthood \cite{papadimitriou-etal-2021-deep}, among others. Sentence embeddings have also been probed using classifiers, and determined to encode specific types of linguistic information, such as subject-verb agreement \cite{goldberg2019}, word order, tree depth, constituent information \citep{conneau-etal-2018-cram}, auxiliaries\citep{adi-etal-2017-finegrained} and argument structure \citep{wilson-etal-2023-abstract}. 

Generative models like LLAMA seem to use English as the latent language in the middle layers \cite{wendler-etal-2024-llamas}, while other analyses of internal model parameters has lead to uncovering language agnostic and language specific networks of parameters \cite{tang-etal-2024-language}, or neurons encoding cross-language number agreement information across several internal layers \cite{varda-marelli-2023-data}. It has also been shown that subject-verb agreement information is not shared by BiLSTM models \cite{dhar-bisazza-2021-understanding} or multilingual BERT \cite{mueller-etal-2020-cross}. Testing the degree to which word/sentence embeddings are multilingual has usually been done using a classification probe, for tasks like NER, POS tagging \cite{pires-etal-2019-multilingual}, language identification \cite{winata-etal-2021-language}, or more complex tasks like question answering and sentence retrieval \cite{hu-etal-2020-xtreme}. There are contradictory results on various cross-lingual model transfers, some of which can be explained by factors such as domain and size of training data, typological closeness of languages \cite{philippy-etal-2023-towards}, or by the power of the classification probes. Generative or classification probes do not provide insights into whether the pretrained model finds deeper regularities and encodes abstract structures, or the predictions are based on shallower features that the probe used assembles for the specific test it is used for \citep{lenci2023NLU,hewitt-liang-2019-designing}.

We aimed to answer this question by using a multilingual setup, and a simple syntactic structure detection task in an indirectly supervised setting. The datasets used -- in English, French, Italian and Romanian -- are (approximately) lexically  parallel, and are parallel in syntactic structure. The property of interest is grammatical number, and the task is subject-verb agreement. The languages chosen share commonalities -- French, Italian and Romanian are all Romance languages, English and French share much lexical material -- but there are also differences: French and Italian use a similar manner to encode grammatical number, mainly through articles that can also signal phrase boundaries. English has a very limited form of nominal plural morphology, but determiners are useful for signaling phrase boundaries. In Romanian, number is expressed through inflection, suffixation and case, and articles are also often expressed through specific suffixes, thus overt phrase boundaries are less common than in French, Italian and English. These commonalities and differences help us interpret the results, and provide clues on how the targeted syntactic information is encoded.

Previous experiments have shown that syntactic information -- chunk sequences and their properties -- can be accessed in transformer-based pretrained sentence embeddings \cite{nastase2024identifiable}. In this multilingual setup, we test whether this information has been identified based on language-specific shallow features, or whether the system has uncovered and encoded more abstract structures. 

The low rate of transfer for the monolingual training setup and the decreased performance for the multilingual training setup for both our experimental configurations indicate that the chunk sequence information is language specific and is assembled by the system based on shallow features. Further clues come from the fact that the only transfer happens between French and Italian, which encode phrases and grammatical number in a very similar manner. Embedding the sentence structure detection into a larger system, where it receives an additional learning signal (shared across languages) does not help to push towards finding a shared sentence representation space that encodes in a uniform manner the sentence structure shared across languages. 

\section{Conclusions}

We have aimed to add some evidence to the question {\it How do state-of-the-art systems $\ll$know$\gg$ what they $\ll$know$\gg$?} \citep{lenci2023NLU} by projecting the subject-verb agreement problem in a multilingual space. We chose languages that share syntactic structures, and have particular differences that can provide clues about whether the models learned rely on shallower indicators, or the pretrained models encode deeper knowledge. Our experiments show that pretrained language models do not encode abstract syntactic structures, but rather this information is assembled "upon request" -- by the probe or task -- based on language-specific indicators. 
Understanding how information is encoded in large language models can help determine the next necessary step towards making language models truly deep. 

\paragraph{Acknowledgments}

We gratefully acknowledge the partial support of this work by the Swiss National Science Foundation, through grant SNF Advanced grant  TMAG-1\_209426 to PM. 

\bibliographystyle{acl}
\bibliography{bibliography,custom,anthology_since2015}
\onecolumn
\appendix

\section{Generating data from a seed file}

To build the sentence data, we use a seed file that was used to generate the subject-verb agreement data. A seed, consisting of noun, prepositional and verb phrases with different grammatical numbers, can be combined to build sentences consisting of different sequences of such chunks. Table \ref{tab:data_samples} includes a partial line from the seed file. To produce the data in the 4 languages, we translate the seed file, from which the sentences and BLM data are then constructed.

\newcommand\x{1.5cm}
\newcommand\y{1.7cm}

\begin{table}[h]
  \footnotesize
  \renewcommand{\tabcolsep}{1mm}
    \centering
    \begin{tabular}{p{\x}p{\x}p{\x}p{\y}p{\y}p{\y}p{\y}p{\y}} \hline
    Subj\_sg	& Subj\_pl & P1\_sg	 & P1\_pl & P2\_sg	& P2\_pl &	V\_sg &	V\_pl \\ \hline
   The computer & The computers & with the program	& with the programs &	of the experiment & of the experiments & is broken & are broken	\\ \hline \hline
    \multicolumn{3}{l}{
      \begingroup
       \renewcommand{\arraystretch}{1.5}
       \begin{tabular}{p{3.2cm}p{1.3cm}} \\
       \multicolumn{2}{l}{\bf Sent. with different chunks} \\ \hline
        The computer is broken. & np-s vp-s \\
        The computers are broken. & np-p vp-p \\
        The computer with the program is broken. & np-s pp1-s vp-s \\
        ... & ... \\
        The computers with the programs of the experiments are broken. & np-p pp1-p pp2-p vp-p \\ \hline \vspace{2mm}
       \end{tabular}
       \endgroup
       } 
    & \multicolumn{5}{r}{
     \begingroup
     \renewcommand{\arraystretch}{1.2}
      \begin{tabular}{p{8.5cm}}
         {\bf a BLM instance} \\ \hline
         \ul{Context:} \\
         The computer with the program is broken. \\
         The computers with the program are broken. \\
         The computer with the programs is broken. \\
         The computers with the programs are broken. \\
         The computer with the program of the experiment is broken. \\
         The computers with the program of the experiment are broken. \\
         The computer with the programs of the experiment is broken. \\ \hline
         \ul{Answer set:} \\
         {\it The computers with the programs of the experiment are broken.} \\
        The computers with the programs of the experiments are broken. \\
        The computers with the program of the experiment are broken. \\
        The computers with the program of the experiment is broken. \\
        ... \\ \hline \vspace{2mm}
      \end{tabular}
    \endgroup
    } \\ \hline
    \end{tabular}
    \caption{A line from the seed file on top, and a set of individual sentences built from it, as well as one BLM instance.}
    \label{tab:data_samples}
\end{table}

\clearpage
\section{Example of data for the agreement BLM}
\label{app:agr}

\subsection{Example of BLM instances (type I) in different languages}

\begin{figure*}[!h]
\centering
\footnotesize
\setlength{\tabcolsep}{5pt} 
\begin{tabular}{|p{0.1cm}|p{0.38\textwidth}|} 
\hline
\multicolumn{2}{|c|}{\sc English - Context} \\
\hline
1 & The owner of the parrot is coming. \\
2 & The owners of the parrot are coming. \\
3 & The owner of the parrots is coming. \\
4 & The owners of the parrots are coming. \\
5 & The owner of the parrot in the tree is coming. \\
6 & The owners of the parrot in the tree are coming. \\
7 & The owner of the parrots in the tree is coming. \\
? & ??? \\ \hline 
\end{tabular} 
\begin{tabular}{|p{0.1cm}|p{0.48\textwidth}|} 
\hline
\multicolumn{2}{|c|}{\sc English - Answers} \\
\hline
1 & The owners of the parrots in the tree are coming. \\
2 & The owners of the parrots in the trees are coming. \\
3 & The owner of the parrots in the tree is coming. \\
4 & The owners of the parrots in the tree are coming. \\
5 & The owners of the parrot in the tree are coming. \\
6 & The owners of the parrots in the trees are coming. \\
7 & The owners of the parrots and the trees are coming. \\
? & The owners of the parrots in the tree in the gardens are coming. \\
\hline
\end{tabular}
\end{figure*}

\begin{figure*}[!h]
\centering
\footnotesize
\setlength{\tabcolsep}{5pt} 
\begin{tabular}{|p{0.1cm}|p{0.38\textwidth}|} 
\hline
\multicolumn{2}{|c|}{\sc French - Context} \\
\hline
1 & Le proprietaire du perroquet viendra. \\
2 & Les proprietaires du perroquet viendront. \\
3 & Le proprietaire des perroquets viendra. \\
4 & Les proprietaires des perroquets viendront. \\
5 & Le proprietaire du perroquet dans l'arbre viendra. \\
6 & Les proprietaires du perroquet dans l'arbre viendront. \\
7 & Le proprietaire des perroquets dans l'arbre viendra. \\
? & ??? \\ \hline 
\end{tabular} 
\begin{tabular}{|p{0.1cm}|p{0.48\textwidth}|} 
\hline
\multicolumn{2}{|c|}{\sc French - Answers} \\
\hline
1 & Les proprietaires des perroquets dans l'arbre viendront. \\
2 & Les proprietaires des perroquets dans les arbres viendront. \\
3 & Le proprietaire des perroquets dans l'arbre viendra. \\
4 & Les proprietaires des perroquets dans l'arbre viendront. \\
5 & Les proprietaires du perroquet dans l'arbre viendront. \\
6 & Les proprietaires des perroquets dans les arbres viendront. \\
7 & Les proprietaires des perroquets et les arbres viendront. \\
? & Les proprietaires des perroquets dans l'arbre des jardins viendront. \\
\hline
\end{tabular}
\end{figure*}

\begin{figure*}[!h]
\centering
\footnotesize
\setlength{\tabcolsep}{5pt} 
\begin{tabular}{|p{0.1cm}|p{0.38\textwidth}|} 
\hline
\multicolumn{2}{|c|}{\sc Italian - Context} \\
\hline
1 & Il padrone del pappagallo arriverà. \\
2 & I padroni del pappagallo arriveranno. \\
3 & Il padrone dei pappagalli arriverà. \\
4 & I padroni dei pappagalli arriveranno. \\
5 & Il padrone del pappagallo sull'albero arriverà. \\
6 & I padroni del pappagallo sull'albero arriveranno. \\
7 & Il padrone dei pappagalli sull'albero arriverà. \\
? & ??? \\ \hline 
\end{tabular} 
\begin{tabular}{|p{0.1cm}|p{0.48\textwidth}|} 
\hline
\multicolumn{2}{|c|}{\sc Italian - Answers} \\
\hline
1 & I padroni dei pappagalli sull'albero arriveranno. \\
2 & I padroni dei pappagalli sugli alberi arriveranno. \\
3 & Il padrone dei pappagalli sull'albero arriverà. \\
4 & I padroni dei pappagalli sull'albero arriveranno. \\
5 & I padroni del pappagallo sull'albero arriveranno. \\
6 & I padroni dei pappagalli sugli alberi arriveranno. \\
7 & I padroni dei pappagalli e gli alberi arriveranno. \\
? & I padroni dei pappagalli sull'albero dei giardini arriveranno. \\
\hline
\end{tabular}
\end{figure*}

\begin{figure*}[!h]
\centering
\footnotesize
\setlength{\tabcolsep}{5pt} 
\begin{tabular}{|p{0.1cm}|p{0.38\textwidth}|} 
\hline
\multicolumn{2}{|c|}{\sc Romanian - Context} \\
\hline
1 & Posesorul papagalului va veni. \\
2 & Posesorii papagalului vor veni. \\
3 & Posesorul papagalilor va veni. \\
4 & Posesorii papagalilor vor veni. \\
5 & Posesorul papagalului din copac va veni. \\
6 & Posesorii papagalului din copac vor veni. \\
7 & Posesorul papagalilor din copac va veni. \\
? & ??? \\ \hline 
\end{tabular} 
\begin{tabular}{|p{0.1cm}|p{0.48\textwidth}|} 
\hline
\multicolumn{2}{|c|}{\sc Romanian - Answers} \\
\hline
1 & Posesorii papagalilor din copac vor veni. \\
2 & Posesorii papagalilor din copaci vor veni. \\
3 & Posesorul papagalilor din copac va veni. \\
4 & Posesorii papagalilor din copac vor veni. \\
5 & Posesorii papagalului din copac vor veni. \\
6 & Posesorii papagalilor din copaci vor veni. \\
7 & Posesorii papagalilor și copacii vor veni. \\
? & Posesorii papagalilor din copac din grădini vor veni. \\
\hline
\end{tabular}

\caption{Parallel examples of a type I data instance in English, French, Italian and Romanian}
\end{figure*}

\newpage
\section{Results}
\label{app:results}

\subsection{Chunk sequence detection in sentences}

\begin{table}[h]
    \centering
    \begin{tabular}{l|cccc}
\diagbox{\bf train on}{\bf test on} & EN & FR & IT & RO \\ \hline  
MultiLang & 0.780 (0.039) & 0.865 (0.036) & 0.811 (0.012) & 0.432 (0.025) \\ \hline
EN & {\bf 0.975 (0.008)} & 0.160 (0.005) & 0.141 (0.011) & 0.144 (0.006)\\
FR & 0.207 (0.018) & {\bf 0.978 (0.008)} & 0.206 (0.016) & 0.150 (0.010)\\
IT & 0.179 (0.029) & 0.372 (0.016) & {\bf 0.982 (0.008)} & 0.161 (0.007)\\
RO & 0.164 (0.004) & 0.197 (0.021) & 0.192 (0.011) & {\bf 0.673 (0.038)}\\ \hline
\end{tabular}
    \caption{Average F1 scores (standard deviation) for sentence chunk detection in sentences}
    \label{tab:sentdet}
\end{table}

\subsection{Results on the BLM Agr* data}

\begin{table}[h!]
    \centering
    \begin{tabular}{l|cccc}
\diagbox{\bf train on}{\bf test on} & type\_I\_EN & type\_I\_FR & 
type\_I\_IT & type\_I\_RO \\ \hline  
type\_I & {\bf 0.839 (0.007)} & 0.938 (0.011) & {\bf 0.868 (0.021)} & {\bf 0.462 (0.023)} \\ 
type\_II & 0.696 (0.006) & {\bf 0.944 (0.003)} & 0.759 (0.004) & 0.409 (0.031) \\ 
type\_III & 0.558 (0.013) & 0.791 (0.026) & 0.641 (0.023) & 0.290 (0.027)\\ \hline \hline
 & type\_II\_EN & type\_II\_FR & 
type\_II\_IT & type\_II\_RO \\ \hline 
type\_I &  {\bf 0.748 (0.001)} & {\bf 0.873 (0.006)} & {\bf 0.851 (0.015)} & {\bf 0.448 (0.015)} \\ 
type\_II & 0.642 (0.002) & 0.871 (0.012) & 0.802 (0.002) & 0.394 (0.012)\\ 
type\_III & 0.484 (0.023) & 0.760 (0.027) & 0.691 (0.023) & 0.299 (0.010)\\ \hline \hline
  & type\_III\_EN & type\_III\_FR & type\_III\_IT & type\_III\_RO \\ \hline 
type\_I & {\bf 0.643 (0.003)} & 0.768 (0.004) & {\bf 0.696 (0.022)} & 0.236 (0.004)\\ 
type\_II & 0.585 (0.010) & {\bf 0.797 (0.008)} & 0.693 (0.009) & 0.240 (0.006)\\ 
type\_III & 0.480 (0.026) & 0.739 (0.027) & 0.691 (0.017) & {\bf 0.262 (0.002)}\\ \hline
\end{tabular}
\caption{Multilingual learning results for the BLM agreement task in terms of average F1 over three runs, and standard deviation.}
\label{tab:multilingualBLM}
\end{table}

\begin{table}[h!]
    \centering
    \begin{tabular}{l|cccc}
\diagbox{\bf test on}{\bf train on} & type\_I\_EN & type\_I\_FR & type\_I\_IT & type\_I\_RO \\ \hline  
type\_I\_EN & {\bf 0.884 (0.002)} & 0.123 (0.032) & 0.125 (0.046) & 0.106 (0.034) \\
type\_I\_FR & 0.103 (0.032) & {\bf 0.948 (0.009)} & 0.466 (0.010) & 0.164 (0.029) \\
type\_I\_IT & 0.113 (0.033) & 0.341 (0.018) & {\bf 0.845 (0.010)} & 0.183 (0.021) \\
type\_I\_RO & 0.113 (0.026) & 0.186 (0.014) & 0.188 (0.015) & {\bf 0.733 (0.027)} \\ \hline
type\_II\_EN & {\bf 0.757 (0.015)} & 0.119 (0.009) & 0.129 (0.029) & 0.103 (0.019) \\
type\_II\_FR & 0.132 (0.024) & {\bf 0.868 (0.010)} & 0.433 (0.008) & 0.187 (0.011) \\
type\_II\_IT & 0.100 (0.020) & 0.386 (0.016) & {\bf 0.875 (0.004)} & 0.196 (0.009) \\
type\_II\_RO & 0.088 (0.007) & 0.174 (0.005) & 0.173 (0.006) & {\bf 0.726 (0.009)} \\ \hline
type\_III\_EN & {\bf 0.638 (0.025)} & 0.117 (0.007) & 0.129 (0.028) & 0.108 (0.013) \\
type\_III\_FR & 0.114 (0.007) & {\bf 0.820 (0.013)} & 0.406 (0.013) & 0.169 (0.017) \\
type\_III\_IT & 0.091 (0.009) & 0.337 (0.016) & {\bf 0.806 (0.009)} & 0.170 (0.013) \\
type\_III\_RO & 0.086 (0.008) & 0.170 (0.007) & 0.174 (0.003) & {\bf 0.314 (0.010)} \\ \hline \hline
 & type\_II\_EN & type\_II\_FR & type\_II\_IT & type\_II\_RO \\ \hline
type\_I\_EN & {\bf 0.772 (0.030)} & 0.154 (0.023) & 0.103 (0.014) & 0.090 (0.007) \\
type\_I\_FR & 0.151 (0.006) & {\bf 0.972 (0.006)} & 0.484 (0.015) & 0.143 (0.018) \\
type\_I\_IT & 0.106 (0.014) & 0.417 (0.018) & {\bf 0.791 (0.004)} & 0.151 (0.034) \\
type\_I\_RO & 0.107 (0.002) & 0.177 (0.020) & 0.170 (0.009) & {\bf 0.625 (0.014)} \\ \hline
type\_II\_EN & {\bf 0.670 (0.002)} & 0.158 (0.015) & 0.106 (0.006) & 0.100 (0.010) \\
type\_II\_FR & 0.188 (0.009) & {\bf 0.903 (0.007)} & 0.434 (0.010) & 0.146 (0.013) \\
type\_II\_IT & 0.100 (0.010) & 0.448 (0.011) & {\bf 0.840 (0.003)} & 0.152 (0.020) \\
type\_II\_RO & 0.093 (0.013) & 0.182 (0.008) & 0.159 (0.011) & {\bf 0.636 (0.006)} \\ \hline
type\_III\_EN & {\bf 0.620 (0.005)} & 0.150 (0.012) & 0.116 (0.007) & 0.092 (0.009) \\
type\_III\_FR & 0.168 (0.007) & {\bf 0.870 (0.005)} & 0.386 (0.008) & 0.127 (0.012) \\
type\_III\_IT & 0.091 (0.005) & 0.387 (0.002) & {\bf 0.770 (0.008)} & 0.132 (0.016) \\
type\_III\_RO & 0.082 (0.014) & 0.175 (0.007) & 0.172 (0.003) & {\bf 0.311 (0.017)} \\ \hline \hline
 & type\_III\_EN & type\_III\_FR & type\_III\_IT & type\_III\_RO \\ \hline
type\_I\_EN & {\bf 0.739 (0.012)} & 0.174 (0.023) & 0.154 (0.013) & 0.059 (0.009) \\
type\_I\_FR & 0.160 (0.007) & {\bf 0.923 (0.013)} & 0.434 (0.005) & 0.196 (0.029) \\
type\_I\_IT & 0.132 (0.011) & 0.384 (0.016) & {\bf 0.797 (0.009)} & 0.197 (0.005) \\
type\_I\_RO & 0.091 (0.011) & 0.164 (0.023) & 0.170 (0.022) & {\bf 0.280 (0.010)} \\ \hline
type\_II\_EN & {\bf 0.662 (0.008)} & 0.164 (0.009) & 0.142 (0.015) & 0.076 (0.010) \\
type\_II\_FR & 0.202 (0.013) & {\bf 0.883 (0.001)} & 0.454 (0.010) & 0.203 (0.010) \\
type\_II\_IT & 0.111 (0.004) & 0.425 (0.005) & {\bf 0.840 (0.002)} & 0.203 (0.006) \\
type\_II\_RO & 0.086 (0.007) & 0.158 (0.006) & 0.158 (0.012) & {\bf 0.379 (0.013)} \\ \hline
type\_III\_EN & {\bf 0.654 (0.010)} & 0.155 (0.006) & 0.140 (0.016) & 0.082 (0.007) \\
type\_III\_FR & 0.183 (0.003) & {\bf 0.860 (0.004)} & 0.431 (0.004) & 0.191 (0.003) \\
type\_III\_IT & 0.106 (0.003) & 0.373 (0.003) & {\bf 0.836 (0.005)} & 0.182 (0.004) \\
type\_III\_RO & 0.082 (0.001) & 0.156 (0.007) & 0.155 (0.007) & {\bf 0.353 (0.006)} \\
\end{tabular}
    \caption{Results as average F1 (sd) over three runs, for the BLM subject-verb agreement task, in the monolingual training setting.}
    \label{tab:monolingualBLM}
\end{table}

\end{document}